\documentclass{llncs}

\usepackage{times,epsfig,graphicx}
\usepackage{algorithm,algorithmic}
\usepackage[usenames]{color}
\usepackage{float}
\usepackage{amssymb,amsmath}
\usepackage[latin1]{inputenc}
\usepackage{fancyhdr}
\usepackage{verbatim}
\usepackage{tabularx}
\usepackage{xspace}
\usepackage{bm}

\newcommand{\hide}[1]{}

\newcommand{\bfx}{{\bf x}}

\title{Mining Smart Card Data for Travelers' Mini Activities}
\author{Boris Chidlovskii}
\institute{Xerox Research Centre Europe \\
F--38240 Meylan, France\\
\email{chidlovskii@xrce.xerox.com}
}

\begin{document}
\date{}
\maketitle
\thispagestyle{empty}

\begin{abstract}
In the context of public transport modeling and simulation, we address the problem of mismatch between 
simulated transit trips and observed ones. We point to the weakness of the current travel demand modeling process;
the trips it generates are over-optimistic and do not reflect the real passenger choices. We introduce the notion of mini
activities the travelers do during the trips; they can explain the deviation of simulated 
trips from the observed trips. We propose to mine the smart card data to extract the mini activities. We develop a technique to integrate them in the generated trips and learn such an integration from two available sources, the trip history and trip planner recommendations. For an input travel demand, we build a Markov chain over the trip collection and apply the Monte Carlo Markov Chain algorithm to integrate mini activities in such a way that the selected characteristics converge to the desired distributions. We test our method in different settings on the passenger trip collection of Nancy, France. We report experimental results demonstrating a very important mismatch reduction.
\end{abstract}

\section{Introduction}
\label{sec:intro}

In many cities around the world, the public transportation systems use smart cards for the trip validation. 
Moreover, the information generated with any card transaction (time and location) represents a rich data source 
for transport and urban planning. Mining data collected by the smart cards allow to observe, understand and model the traveler behaviour on a micro level. 
The {\it travel demand modeling} by mining the smart card data includes several steps, 
such as the trip reconstruction, the origin-destination (OD) extraction, activity inference and micro-simulation~\cite{anda16}.
Implemented in different simulation platforms~\cite{adnan16,Behrisch11,erath14,MAP}, this process proved its efficiency in real public transportation systems~\cite{anda16,balmer09,erath14}.

In this paper, we make a step towards more accurate travel demand modeling. 
We argue that the previous approaches 
lack the capacity to learn from the past data. 
We compare the simulated trips to the observed ones 
and discover an important mismatch between the simulated transit trips and observed ones.
This mismatch concerns all critical characteristics, such as the full trip time, the transit time, modality choices, etc..

We study this phenomenon in detail. While the existing modeling process copes well with the main activities like {\it home}, {\it work}, {\it leisure} and {\it shopping}, we suggest it disregards so called mini activities the travelers do during the trips.
Mini activities are ubiquitous and may include, among others, bringing kids to school, buying a journal or meeting a friend. 
We propose to mine the smart card data for the mini activities and integrate them in the trip generation. We design a method able to learn such an integration from the past observations. Unlike explicit modeling of the main activities, we model the travelers mini activities implicitly. We generate individual trips with integrated mini activities,
by mixing two available sources of trips, the history of observed trips and recommendations proposed by a trip planner. 

Trip planners are a common service available for many cities around the world~\cite{bast16}. 
A planner uses all available network (GTFS) information and service schedules to recommend $k$-top routes upon a user travel request. When recommending a route, it follows one of maximum utility criteria, like "the fastest route", "route with the minimal number of changes", etc. 
These top trip planner recommendations can be directly used for the simulation. Unfortunately they represent an {\it over-optimistic} view of urban traveling and poorly reflect the real passenger route choices~\cite{trepanier05}. 

Another source of information is the history of trips, reconstructed from the smart card data~\cite{mezghani2008}. These trips reflect the route choices users made and various factors that the trip planner may ignore, such as too long transfer delays, non-optimal route choice, etc.. 
When possible, sampling from a trip history is a better strategy for simulating the trips which look like the observed ones. 
However, the history of trips is limited to the time and space where they have been collected. It can not be reused beyond this specific context, for example, when the network is changed or we want to simulate a what-if scenario. 

 
We consider {\it the trip planner recommendations} and {\it trip history sampling} as complementary sources for the {\it realistic trip generation} where simulated trips look like the observed ones. 
We develop a method to model travelers mini activities as a part of the generated trips, by mixing up the trip history and trip planner recommendations, for simulating both existing and new public transport scenarios. 

The realistically looking trips are generated by following the next major steps. We first selects characteristics we want the  generated trip collection to match and express them in the form of desired distributions. 
Then, for an input travel demand, 
we build a Markov chain over the trip collection and apply the Monte Carlo Markov Chain algorithm to integrate mini activities in such a way that the selected characteristics converge to the desired distributions.

The remainder of the paper is organized as follows. Section 2 presents the process of mining smart card data for travel demand modeling. It discusses in detail the problem of mismatch between the simulated and observed trips and then introduces the notion of travelers mini activities. Section 3 recalls the Monte Carlo Markov Chain algorithm and describes its application to integrating the mini activities in the trip generation. Section 4 reports the evaluation results on one real collection of public transport trips.
Section 5 concludes the paper.

\begin{figure}[ht]
\centering{
\includegraphics[width=7.4cm]{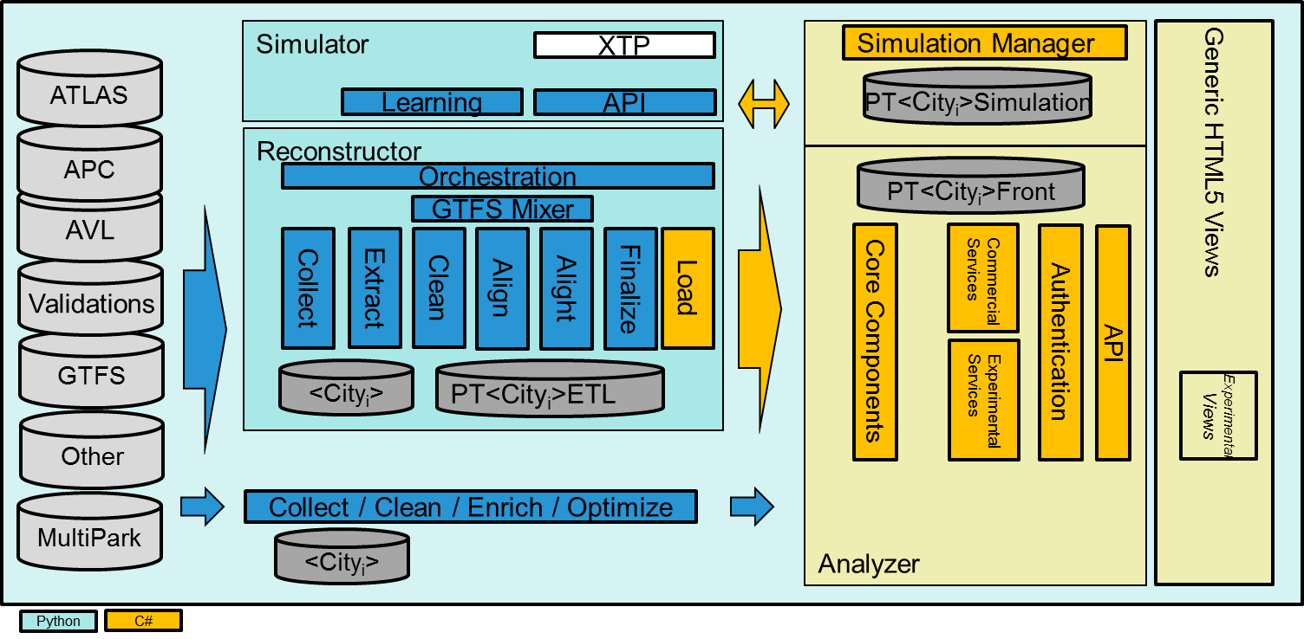}
\includegraphics[width=4.6cm]{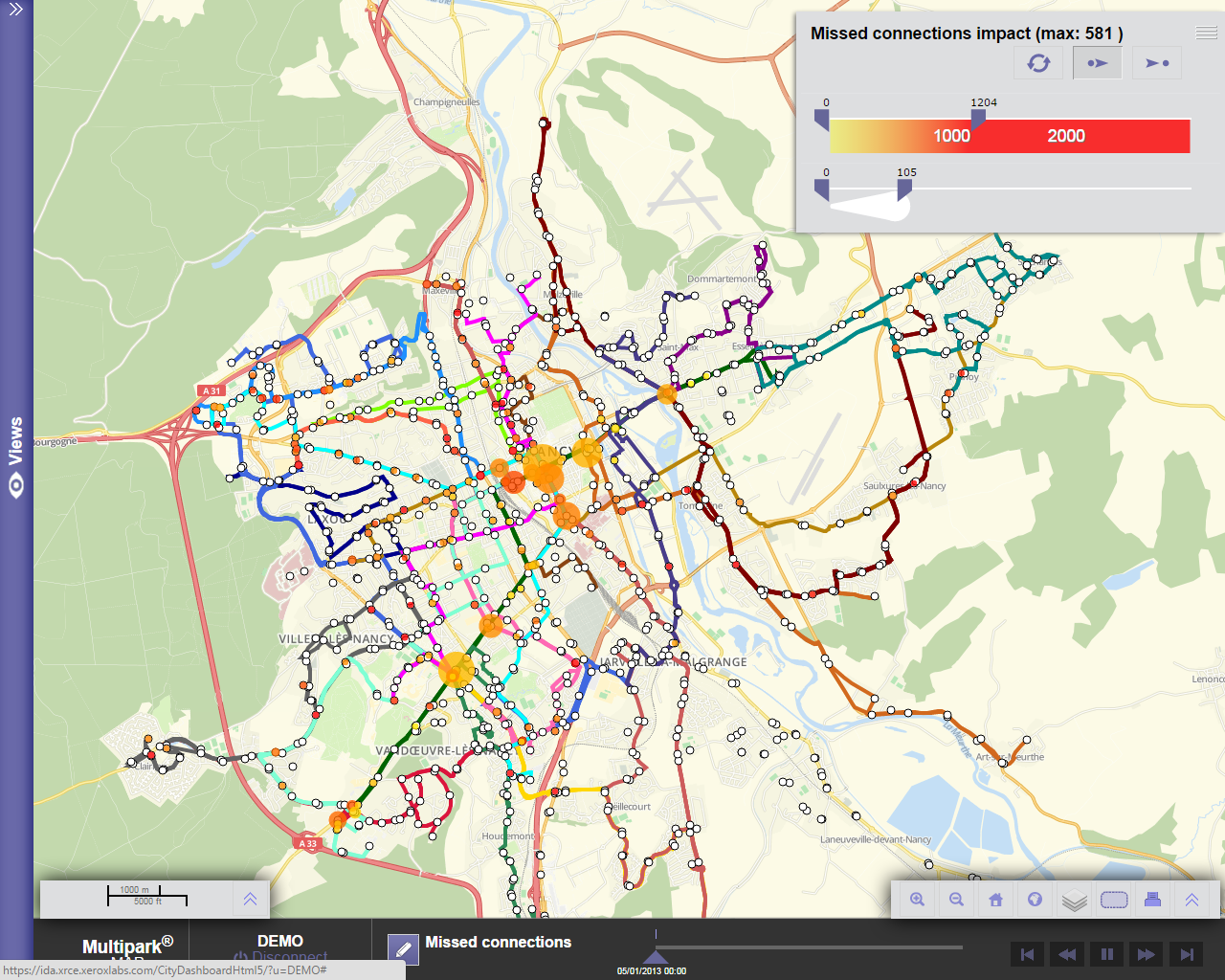}}
\centering{
\includegraphics[width=6.5cm]{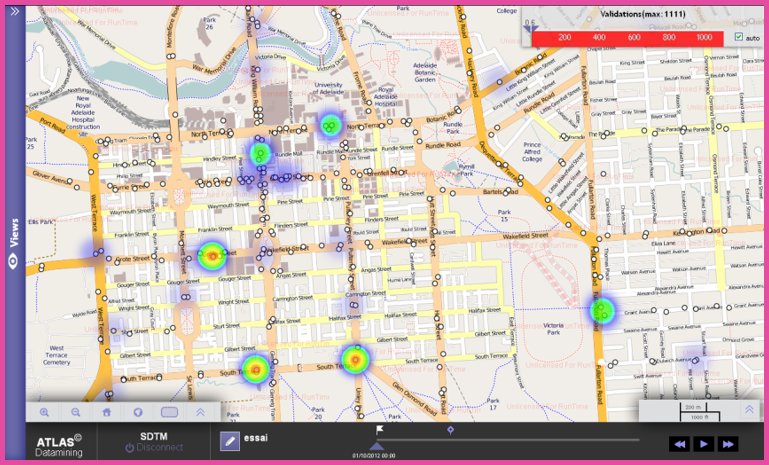}
\includegraphics[width=5.5cm]{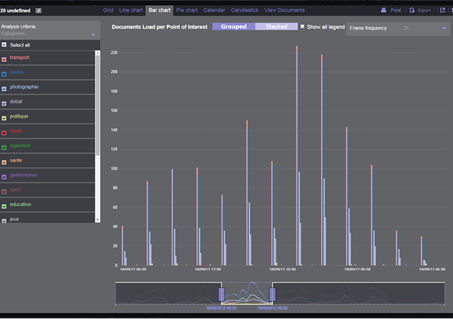}}
\caption{MAP architecture, the visualization module, simulation diagram and analytic interface.}
\label{fig:map}
\end{figure}

\section{Travel demand modeling by mining the smart cart data}
\label{sec:smart}
 
Smart card automated fare collection (AFC) systems are used in public transportation systems all around the world. 
The main objective of using smart cards for public transport is its ability for flexible and secure fare
collection~\cite{agard06}. Moreover, data collected by smart cards (time and location) 
has been recognized as a rich data source for urban planning and transport modeling. 

It is common to recognize the following steps when modeling the travel demand from the smart card data~\cite{anda16,fourie16}:

\begin{enumerate}
\item {\it Individual trip reconstruction}: in transportation systems with a fixed fare, only the boarding validation is required. The first challenge of mining smart card data is to reconstruct the individual trips from the boarding validations.
Different versions of the trip-chaining algorithm~\cite{barry02} are often used; they estimate the alighting stops 
and infer missing links to extract the consistent individual trips. 

\item {\it Origin-Destination (OD) extraction}: once the alighting locations are known, the second step 
 is to infer whether the alighting location is the final destination, and to scale it up to the lost/missing trips. 
Once individual trips are reconstructed with the known boarding locations and
final destinations, public transport OD matrices can be calculated in a straightforward way.

\item {\it Activity modeling}: Public transport individual trips can be further studied to give a semantic meaning to the
inferred locations, like home, work, shopping and leisure. 
By using rule-based or learning-based approaches, activities are identified mainly by the smart card type and temporal attributes of the trips~\cite{agard06,charikov12}.

\item {\it Multi-agent simulation and trip assignment}: smart card data can be further coupled with the network data to reconstruct the bus trajectories. The state of art simulation frameworks, like MATSim~\cite{balmer09}, TRANSIMS\cite{smith95} and SimMobility~\cite{adnan16},
can then generate activity plans for each agent in the simulation, both vehicles and passengers.
Multi agent-based modeling is built upon a large scale of autonomous agents which perform
their own decisions, interact with one another and with the environment. For each agent, an
initial daily activity plan is assigned as a precise description of the activities location, its
durations, start and end time, and the trips connecting two activities.

For example, in MATSim, the day is simulated iteratively and after each iteration a fraction of the agents is
allowed to modify their plans. 
At the end of each simulated day, the utility function is measured for each agent using
a scoring function. 
Agents seek to improve their utility over iterations until the system
reaches an equilibrium where the generalized utility can not be longer improved~\cite{anda16}.
\end{enumerate}

\subsection{MAP}
\label{ssec:map}
Our working framework is the Mobility Analytics Platform (MAP) developed at XRCE\footnote{https://www.news.xerox.com/news/Xerox-Mobility-Analytics-Platform.} and deployed in different cities around the world~\cite{MAP}.
MAP implements all steps of the modeling process described above, exempt the micro-simulation step.
Unlike other platforms, the MAP delegates the candidate generation for trip assignment, the utility evaluation and re-planning to a city trip planner~\cite{bast16}. Using tfast and efficient services of existing trip planners allows to increase the scalability and cope with a very large number of individual agents, both vehicles and travelers, without simplifying the network and scenarios as the MATSim does~\cite{fourie12}.
Figure~\ref{fig:map} presents the architecture, the visualization module, the simulation diagram and the analytic interface of the current MAP platform.

\subsection{Mismatch between simulated and observed trips}
\label{ssec:mismatch}

Like in any computer simulation, the travel demand simulation aims to reproduce the real demand using the inferred model. 
By running the model in a simulator and comparing the simulation results 
to experimental ones, we can gain an insight how to improve the model~\cite{barcelo10}. 
To our best knowledge, this is the first attempt to do such a comparison 

Once all components of a travel demand model are known, the simulator can generate all individual trips for a given day 
and compare them to the trips observed the same day. 
%
The major challenge for all existing simulators is than they fail to accurately reproduce the transit trips. 
Below we report the results of running the MAP simulator using a collection of reconstructed trips for the city of Nancy, France. 
When simulating the trips for a given day $d$ in the collection, with the model inferred from a set of 
reconstructed trips, it {\it systematically} deviates from the trips observed the same day. 

\begin{figure}[ht]
\centering{
\includegraphics[width=8cm]{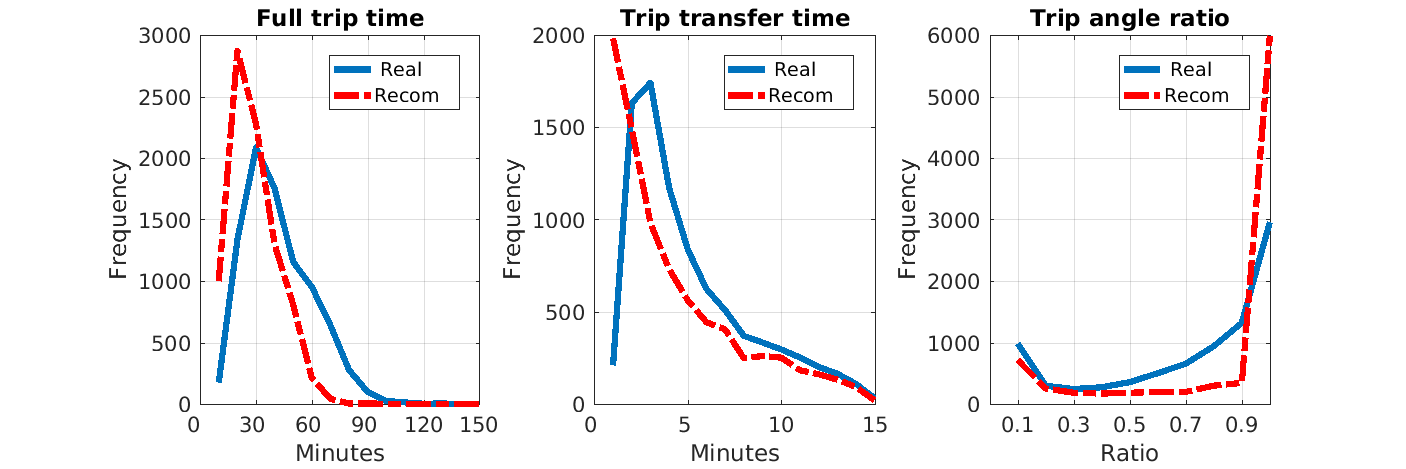}
\includegraphics[width=4cm]{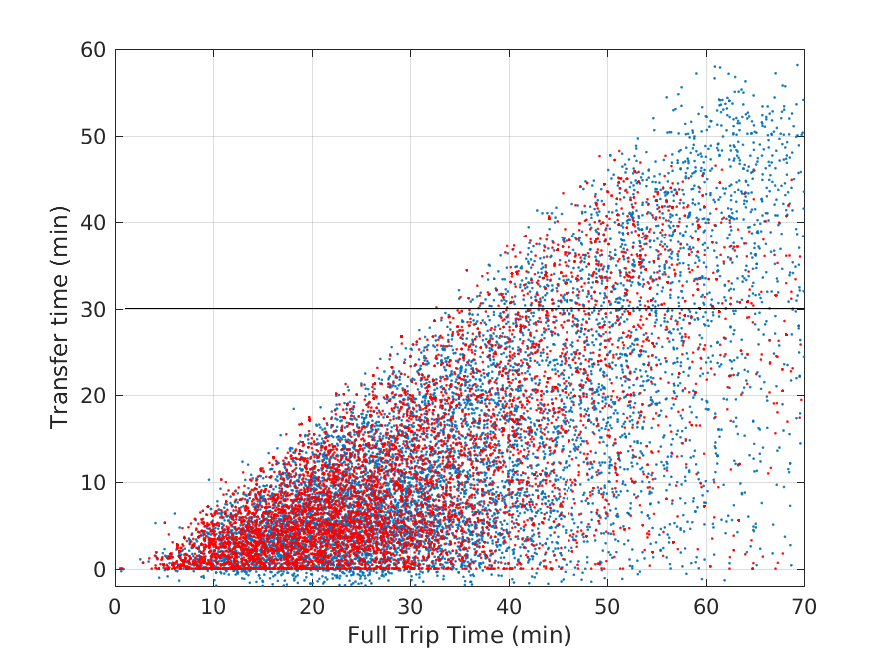}
}
\caption{Left) Mismatch between the simulated (red) and observed (blue) trips; 
         right) full vs transfer time distribution, with the 30 min threshold.}
\label{fig:mismatch}
\end{figure}

Figure~\ref{fig:mismatch} shows this mismatch in details. We consider two standard characteristics of a trip, the {\it full trip time} and the {\it transit trip time}. The former refers to the time between the first boarding and the last alighting, the latter is the time spent when changing/waiting a bus, walking between stops, etc. Both values are readily available for all reconstructed trips. For a usual day in the Nancy collection, the simulated trips have the full trip time on average 36\% shorter (35 mins instead of 23) that the observed trips; the transfer time is 35\% shorter as well (9 mins instead of 14). 
Figure~\ref{fig:mismatch} compares the distributions of these two characteristics, as well as the {\it trip angle ratio} (see Section~\ref{sec:eval} for more detail) of the observed and simulated trips.

It is obvious that the simulated trips are {\it over-optimistic}, they underestimate some factors of travelers choices.
There may exist different explanations to this.
First, the OD extraction step can be put into question. Indeed, 
trips are often split into segments using a threshold, like 30 minutes per activity~\cite{charikov12}. 
Obviously such an approach overlooks any shorter activity; moreover tuning the threshold value would not solve the problem. 
Figure~\ref{fig:mismatch}.right) reports the joint 2D distribution of full trip time and the transfer time for for simulated (blue) and observed (red) trips; it also shows the 30 minute threshold. Looking at the difference between the blue and red clouds, no threshold value can correctly fix the mismatch between the two.

In this paper we propose another explanation to the discovered mismatch; it concerns the activity inference of the demand modeling process. Conventionally, it considers trips connecting main human activities: home, work, leisure and shopping.
We argue that, beyond these main activities, travelers do other things {\it during} the trips. These, so called {\it mini activities} last minutes but cause deviations from the optimal trip plans, in both the time and route choice. 
Consequently, it misleads the maximum utility used for the trip assignment by all existing simulators.

\begin{figure}[ht]
\centering{
\includegraphics[width=6.05cm]{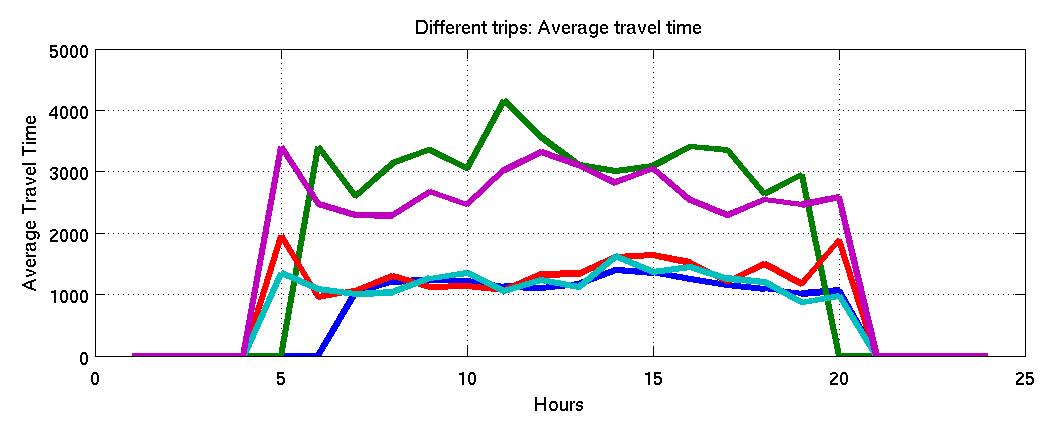}
\includegraphics[width=6.05cm]{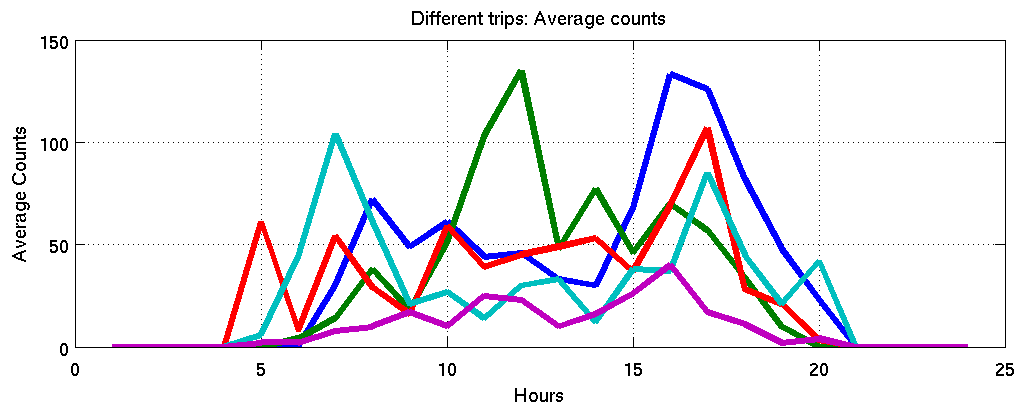}
}
\caption{5-top trips for one OD pair in Nancy: a) the full travel time during the day, b) trip counts.} 
\label{fig:example}
\end{figure}

Figure~\ref{fig:example}.a shows an example; it reports 5-top transit trips for one OD pair in Nancy city. 
It plots the full trip time and trip counts during the day. 
The trip planner, which implements the maximum utility criteria, recommends the trips shown in red, blue and magenta. 
Two other trips, shown in pink and green, are much slower and never recommended despite being frequently used. Moreover, the green one dominates all others at lunch time, possibly due to the closeness to a shopping center. 

\subsection{Mini activities}
\label{ssec:micro}
The mismatch between the simulated and observed trips has been mostly disregarded in the previous approaches to the 
travel demand modeling. Mini activities are numerous; everyone can name bringing kids to school, buying a journal, meeting a friend, taking a walk through a park, and many others. Due to their variability, it looks impossible to name or to even enumerate them. The impact of any specific mini-activity looks negligible, but their total number leads to a sensible deviation from what is considered as the maximum utility or optimal trip plans. An extreme example is {\it one ticket round trips}, very popular in some cities and within some categories of travelers. 
In Nancy, this category represents up to 18\% of individual trips on Tram 1, the highest density service in the city.

Mini activities are hard to recognize, they are integrated in the trips and can cause deviations from the optimal trip plans. The percentage of trips deviated by the mini activities is estimated up to 45\% of transit trips. Instead of enumerating the micro activities and explicitly tuning the utility function, we propose a method that processes them implicitly.


We develop a method to generate trips with integrated travelers mini activities, in such a way they look like the observed trips (see Figure~\ref{fig:mismatch}). It goes beyond the travelers main activities (home, work, leisure, shopping)
and naturally integrates micro-activities in the route assignment. 


The trips should be generated in such a way that their characteristics follow the desired distributions. These distributions can be either empirical (see three histograms in Figure 2) or defined as the distribution functions. For example, the total trip time can be approximated by a mixture of Gaussians, the transit time closely follows the Poisson distribution and the trip angle ratio is approximated by two-mode Beta distributions.

 


In the next section, we describe our method for generating trips with the integrated mini activities. 
It builds a Markov chain over the trip collection, initiates it , and applies the Monte Carlo Markov Chain algorithm in such a way that the main characteristics converge to the desired distributions.
It changes the trips by sampling from two available trip sources, the trip planner recommendations and the trip history.


\section{Markov Chain Monte Carlo for trip generation}
\label{sec:model}
Markov Chain Monte Carlo (MCMC) methods is a class of algorithms for sampling from a probability distribution; they construct a Markov chain that has the desired distribution as its equilibrium distribution. The state of the chain is used as a sample of the desired distribution and the quality of the sample improves over the number of steps. The Metropolis-Hastings (MH) version of MCMC generates a random walk using a {\it proposal density} and a method for {\it rejecting} some of the proposed moves~\cite{andrieu03}.

Given a target probability density $p$, defined on a state space $\bfx$ and computable up to a multiplying constant, the standard MH algorithm proposes a generic way to construct a Markov chain on $\bfx$ that is 
stationary with respect to $p$. It means that if $\bfx^{(i)} \sim p(\bfx)$, then $\bfx^{(i+1)}\sim p(\bfx)$.
The Markov chain generated by the method, $\bfx^{(1)},\bfx^{(2)},\ldots,\bfx^{(t)},\ldots$ is such that $\bfx^{(t)}$ converges {\it in distribution} to $p$. 

In our case, any state $\bfx^{(i)}$ of the Markov chain is an instance in a multi-dimensional space of route assignments for the individual trips. We are interested in such a version of MH that does not sample, but minimizes the target function $p$. As the MH algorithm might become inefficient for the optimization due to exploration of vast areas of no interest, we extend it with the {simulated annealing} (SA) component~\cite{andrieu03} which is a decreasing cooling mechanism for function $p(\bfx)$, in the form of $p^{1/L_i}(\bfx)$ where $lim_{i \rightarrow \infty} L_i = 0$. 
The algorithm is described below.

\begin{algorithm}[h]
\begin{algorithmic}[1]

\STATE{Initialize $\bfx^{(0)} \sim q(\bfx)$; $L_0=1$; }
\FOR {iteration $i = 0,1,2,\ldots$} 
  \STATE {Propose a candidate $\bfx^{cand} \sim q(\bfx^{(i+1)}|\bfx^{(i)})$}
  \STATE {Acceptance probability
     $\alpha(\bfx^{cand}|\bfx^{(i)})={\rm min}\{1, 
     \frac{q(\bfx^{(i)}|\bfx^{cand}) p^{1/L_i}(\bfx^{cand})}
          {q(\bfx^{cand}|\bfx^{(i)}) p^{1/L_i}(\bfx^{(i)})}\}$}
  \STATE {$u \sim {\rm Uniform} (u,0,1)$}
  \IF {$u < \alpha$}
    \STATE {Accept the proposal: $\bfx^{(i+1)} \leftarrow \bfx^{cand}$}
  \ELSE
    \STATE {Reject the proposal: $\bfx^{(i+1)} \leftarrow \bfx^{(i)}$}
  \ENDIF
  \STATE {Set $L_{i+1}$ according to a cooling procedure, for example, $L_{i+1}=0.99 \cdot L_{i}$.}
\ENDFOR
\RETURN {$\bfx$} 

\end{algorithmic}
\caption{Metropolis-Hastings algorithm with simulated annealing (MH-SA).}
\label{alg-mcmc}
\end{algorithm}

The first step of the MH-SA algorithm is to initialize all variables in $\bfx$, by sampling from the candidate set $C(T)$.
The main loop of the algorithm includes three steps. First, it generates a candidate from the proposal distribution $q(\bfx^{(i+1)}|\bfx^{(i)})$. Second, it computes the acceptance probability via the acceptance function $\alpha(\bfx^{cand}|\bfx^{(i)})$ based on the proposal distribution $q$, the target function $p$ and the cooling coefficient $L_i$. Third, it accepts the candidate with probability $\alpha$, or rejects it with probability $1-\alpha$.

\subsection{Mixed sampling for the trip generation}
\label{ssec:appli}
We discuss now how Algorithm 1 applies to the trip generation. As input, we have the OD matrix defined by a set of triples $t_j$, $T=\{t_1,\ldots,t_n\}$, where each triple ($o$,$d$,$t_s$) is a demand to travel from origin $o$ to destination $d$ at time $t_s$.
For a given triple $t_j$, let $C(t_j)$ denote a set of possible route assignments for $t_j$, (see an example with 5-top trips in Figure~\ref{fig:example}). Let variable $x_j$ denote one realization of $t_j$, $x_j \in C(t_j)$. Put all together, variables $x_j$ form the state vector $\bfx = [x_1,\ldots,x_n]$ of the Markov Chain that we want to build.

For any route assignment $x_j$ of triple $t_j$, we use a characteristic function $f(x_j)$, such as the full trip or transit time, in order to compare the generated trips to the observed ones. Let $Z$ denote the desired distribution for function $f$. We want to find such route assignments $x_j, j=1,\ldots,n$ that the function value set, $f({\bf x})=\{f(x_1),\ldots,f(x_n)\}$, follows the desired distribution $Z$, $f({\bf x}) \sim Z$. If we work with multiple characteristic functions $f_1, f_2, \ldots, f_m$, each $f_k$ should follow a distribution $Z_k$, $k=1,\ldots,m$.

The problem is therefore defined over all possible route assignments in $T$, ${\bf x}\in C(T)$, where $C(T)=C(t_1) \times \ldots \times C(t_n)$. We are looking for such an assignment of variable vector $\bfx^*$ that minimizes the target function $p(\bf x)$,
\begin{equation}
{\bf x}^* = argmin_{{\bf x} \in {\bf x}_1,{\bf x}_2,\ldots}p({\bf x}),
\label{eq:loss}
\end{equation}
where the function $p({\bf x})=\sum_k ||{f_k}({\bf x})-Z_k||_1$ represents the mismatch (error) presented in Section 2.2; it takes its minimum when all $f_k({\bf x})$ fit the corresponding $Z_k$.

\subsection{Proposal distribution}
\label{ssec:prop}
Success of any MCMC algorithm depends on an accurate design of the proposal distribution $q(\bfx^{(i+1)}|\bfx^{(i)})$~\cite{andrieu03}. In our setting, 
we factorize $q$ over all variables in $\bfx$ and allow only one variable $x_j$ to change in $\bfx^{(i)}$, 
by sampling another route assignment from the candidate set $C(t_j)$. In Line 4 of Algorithm 1, the replacement of current 
$x^{(i)}_j$ by another candidate $x^{cand}_j \in C(t_j)$ is reduced to the following:
$$
\frac{ q(\bfx^{cand}|\bfx^{(i)})} {q(\bfx^{(i)}|\bfx^{cand}) } = 
\frac {\prod_k Pr(x_k^{cand})} {\prod_k Pr(x^{(i)}_k)} = \frac {Pr(x_j^{cand})} {Pr(x^{(i)}_j)}.
$$
Foe any $t \in T$, we populate the candidate sets $C(t)$ by using the history of trips and the trip planner recommendations, as follows:
\begin{enumerate}
\item 
$C(t)$ includes all trip planner recommendations for $t$; generally we have no access to the recommendation weights, so we consider them as equi-probable. Then $\frac {Pr(x_j^{cand})} {Pr(x^{(i)}_j)}=1$ and we obtain a simpler Hastings algorithm where 
$\alpha(\bfx^{cand}|\bfx^{(i))})={\rm min}\{ 1, p(\bfx^{cand})/ p(\bfx^{(i)}) \}$.
\item 
$C(t)$ includes, if available, all candidates for $t$ from the trip history. As the trip frequencies are usually available from the history, we take them into account. In this case, $Pr(x_j^{cand}) \neq Pr(x^{(i)}_j)$.
\end{enumerate}



For each travel triple $t \in T$, we compose the candidate set $C(t)$ by merging the $k$-top trip planner recommendations with the trip routes from the history, with their weights. Then we run Algorithm 1 with the state vector $\bfx \in C(T)$, the target function $p(\bfx)$ and the proposal function $q$ as described above.

\section{Evaluation}
\label{sec:eval}

We test the trip generation with integrated micro-activities on
the Nancy transit trip collection. The collection includes 1.2M reconstructed transit trips, extracted from the smart card data collected in Nancy, France, during 4 weeks in 2013. Figure~\ref{fig:nancy}.left shows the number of trips for every day in the collection. Note the difference between the working days and weekends and missing data for three days.

For a given day $d$ in the collection, we set the target distributions $Z_i$ as an average of all trips of the same type (for ex, working days) occurred prior to the day $d$. Three characteristic measures are the following:
\begin{itemize}
\item The full travel time $f_1$, with the desired distribution $Z_1$ (Figure~\ref{fig:mismatch}.a),
\item The transfer time $f_2$, with the desired distribution $Z_2$ (Figure~\ref{fig:mismatch}.b),
\item The trip angle ratio $f_3$, with the desired distribution $Z_3$ (Figures~\ref{fig:mismatch}.c).
\end{itemize}

\begin{figure}[ht]
\centering{
\includegraphics[width=6.6cm]{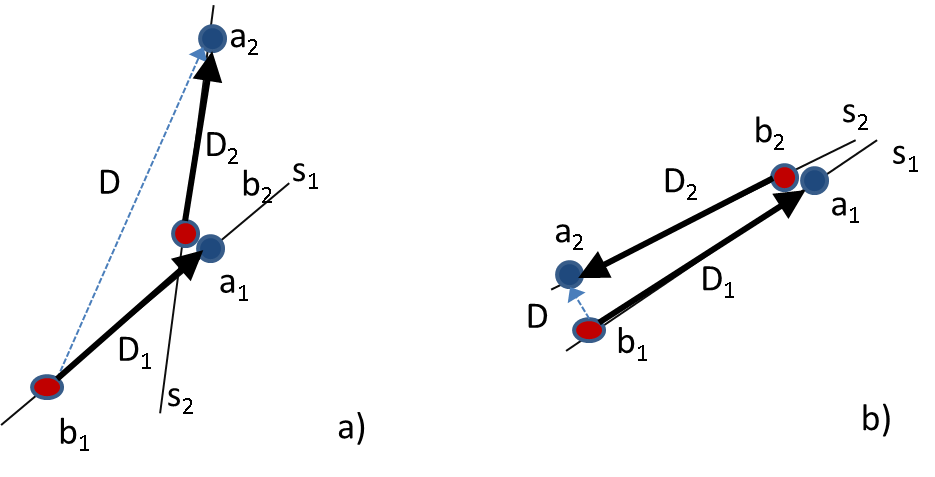}
\includegraphics[width=5.5cm]{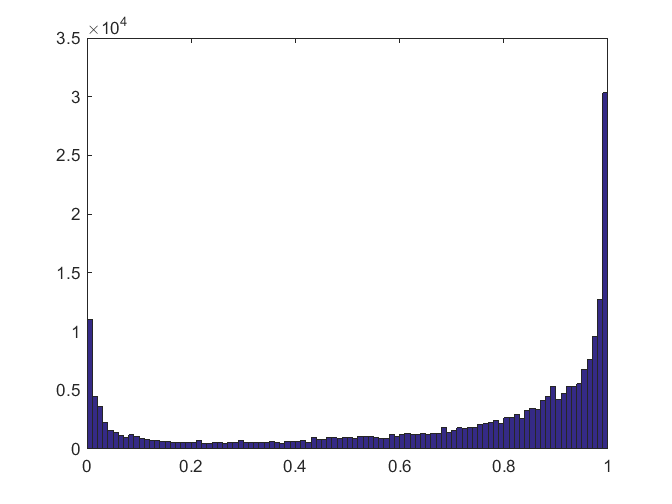}}
\caption{Trip angle examples: a) likely an optimal trip, b) likely a round trip. 
c) Trip angle ratio distribution for Nancy transit trips.}
\label{fig:ratio}
\end{figure}

Measures $f_1$ and $f_2$ are standard trip characteristics. The trip angle ratio $f_3$ aims to distinguish trips looking like optimal plans from non-optimal ones. For a trip $x$ with $m$ legs, defined by a sequence of known boardings $b_i$ and alightings $a_i, i=1,\ldots,m$, the {\it trip ratio} measures the distance $D$ connecting the origin $b_1$ to destination $o_n$, to the sum of leg distances $\gamma=\frac {D}{\sum_{i=1}^n D_i}$. For the sake of geometric interpretation, we approximate this ratio as a tangent and get the corresponding angle normalized to [0,1] range, as follows
\begin{equation}
f_{3}=\frac{2}{\pi} \arctan \frac {D}{\sum_{i=1}^n D_i-D}.
\label{eq:gamma}
\end{equation}
Function $f_{3}$ measures how far the trip is deviated from the theoretically direct connection from origin $o=b_1$ to destination $d=o_n$. Figure~\ref{fig:ratio} demonstrates the idea. For trip planner recommendations, the ratio is close to 1 (Figure~\ref{fig:ratio}.a); for the round trips, the ratio is close to 0 (Figure~\ref{fig:ratio}.b).

It turns out that $f_3$ values reflect well the dichotomy observed in real reconstructed trips, with two modes indicating that trips with angle ratios close 1 and 0 dominate the distribution. Figure~\ref{fig:ratio}.c shows the empirical trip angle distribution for transit trips in Nancy collection. As mentioned before, $f_3$ values can be approximated by the Beta distribution. For example, the empirical distribution in Figure~\ref{fig:ratio}.c fits the Beta distribution with $\alpha=0.26, \beta=0.24$.

\subsection{One day evaluation}
\label{ssec:setting}

\begin{figure}[ht]
\centering{
\includegraphics[width=8.5cm]{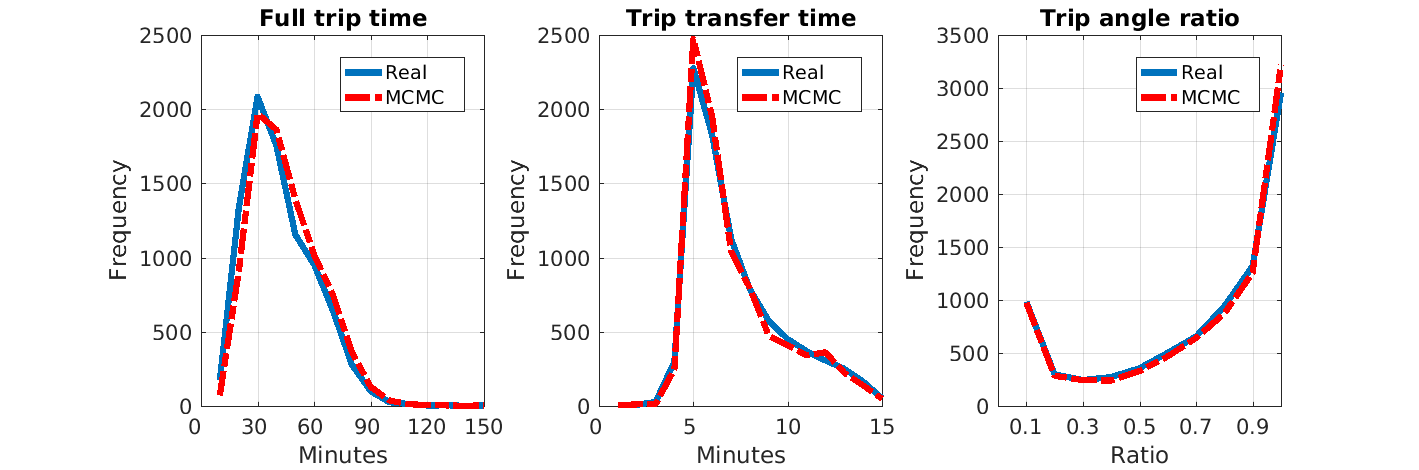}
\includegraphics[width=3.5cm]{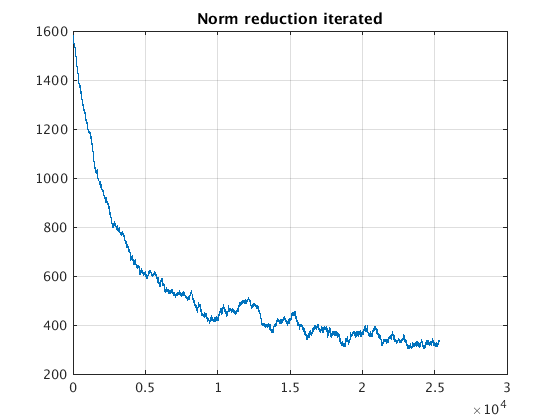}}
\caption{Left) Distribution differences after MH-SA for functions $f_1$, $f_2$, $f_3$; right) Error reduction over iterations.}
\label{fig:mcmc}
\end{figure}

Figure~\ref{fig:mcmc} shows results of applying MH-SA algorithm. For the sake of comparison, we select day $d$=52 (Wednesday), used to demonstrate the mismatch in Figure 2. Target distributions $Z_1$, $Z_2$, $Z_3$ are averaged over all working days prior to the day $d$. 


The MH-SA algorithm converges to the state $\bfx$ which closely satisfies the target distributions.
Figure~\ref{fig:mcmc}.left) shows how the distribution mismatch is reduced for all three characteristic functions. 
We also note a relatively fast convergence of the MH-SA algorithm. 
Figure~\ref{fig:mcmc}.right) shows how the error $p(\bfx)$ decreases as a function of the iteration number. 

\subsection{Error reduction over time}
\label{ssec:time}

We then test the method in the online evaluation setting, where {\it training} trips are those occurred before the selected ({\it test}) day $d$. This implies that the available training data is different for each test day. Also, for early days in the
collection, fewer training data are available. The online evaluation naturally fits the real-world case
in which every new day is used for testing a model trained on all previously observed trips.

We show how the available training history influences the trip generation error. The Nancy collection spreads over 25 days, with 17 working days and 6 weekends. We first run the experiment on the sequence of 17 working days only.
Figure~\ref{fig:mcmc}.left) reports the error for the available training data, after 0, 25K, 50K, 75K and 100K iterations. The error is getting smaller with the longer history, as more relevant trips makes it easier to find candidates to reduce the error.
The generation error is reduced from 0.10 when one day history is available, to 0.03 error when 16 previous days are available.

\begin{figure}[ht]
\centering{
\includegraphics[width=7cm]{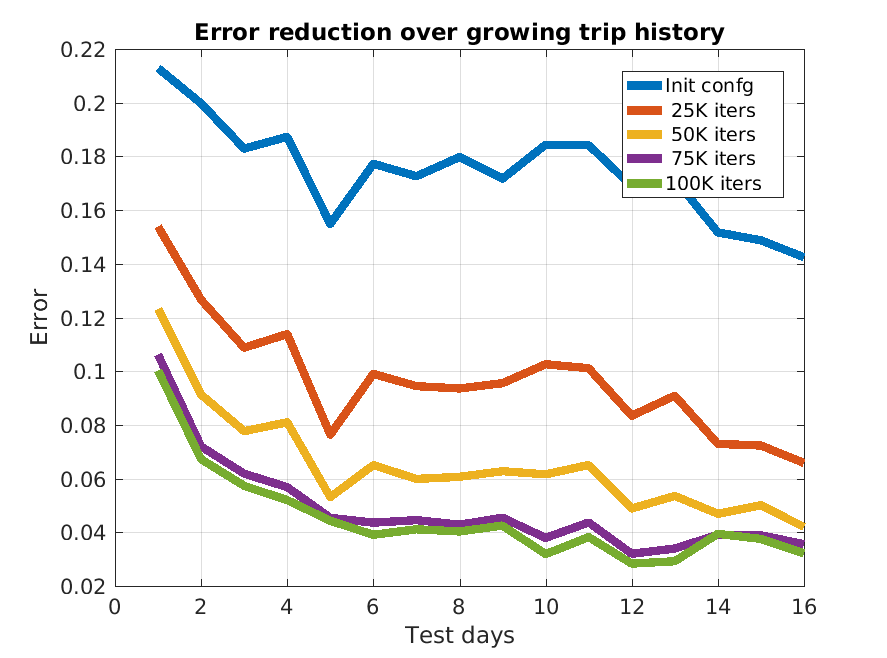}}
\caption{Error reduction over the growing training set (working days only).}
\label{fig:hist}
\end{figure}

\subsubsection{Mixing working days and weekends.}
\label{sssec:exp2} 
\begin{figure}[ht]
\centering{
\includegraphics[width=6cm]{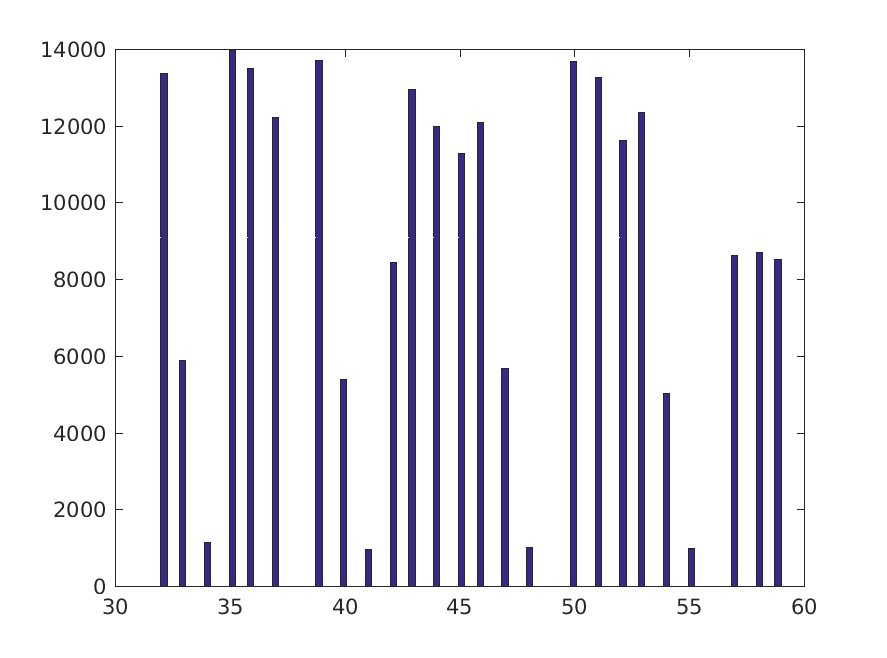}
\includegraphics[width=6cm]{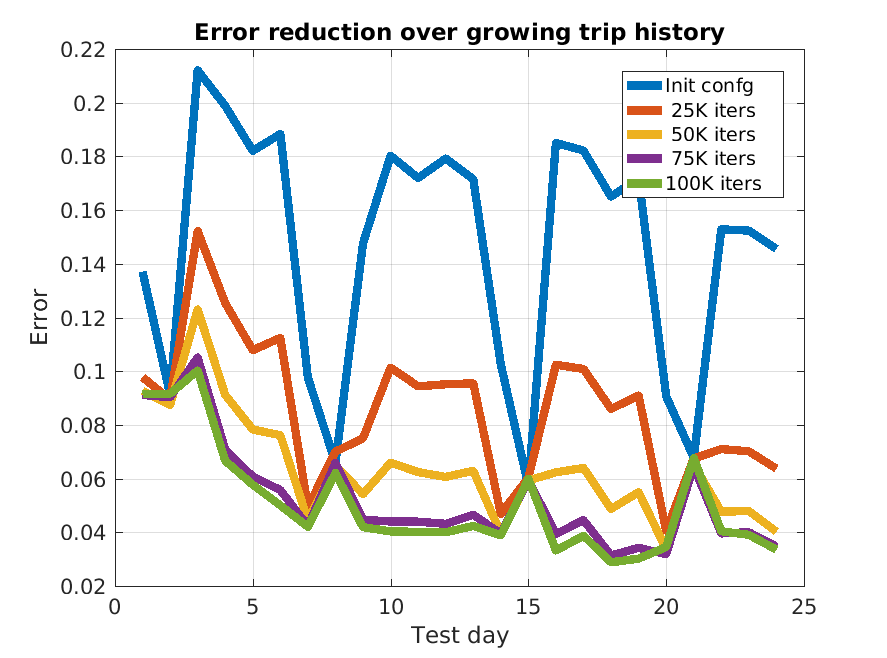} }
\caption{left) Number of reconstructed trips per day in Nancy set. Right) Error reduction with all the days.}
\label{fig:nancy}
\end{figure}

To complete the picture of trip generation, we run the experiment with making no distinction between working days and weekends. Figure~\ref{fig:nancy} reports the generation error for the entire sequence of 25 days, which includes 16 working days, Saturdays (days 33, 40, 47 and 54) and Sundays (days 34, 41, 48 and 55). As the figure shows, trip generation for working days does not profit from available weekends trips and vice versa. This supports the known principle of a split between working days and weekends,
as presenting different traveling patterns. They are therefore hardly mixable when generating trips for either day. 

\section{Conclusion}
\label{sec:conclusion}

In this paper we propose a method for generating trips which integrate travelers mini activities and look realistic, with main characteristics fitting the desired distributions. In the current version, the method is able to reproduce the observed trips, by mixing available trip sources, the trip history and the trip planner recommendations. It generates a Markov chain over the trip collection and uses a version of Metropolis-Hastings algorithm to make it converging to a desired distribution. 
We test our method in different settings on the passenger trip collection of Nancy, France. We report experimental results demonstrating a very important mismatch reduction. The current MAP simulator addresses partially the mismatch by tuning the trip planner parameters, this allows to reduce the mismatch to some degree, but can not solve it completely. Taking into account the travelers mini activities and integrating them in the travel demand modeling helps boost the simulation accuracy. 
As the next step, we target the travel demand modeling where the desired distributions do not necessarily follow the past observations. Such generalized models can be served as an input to simulate different {\it what-if} scenarios, when the history of trips is partial or does not exist at all. 

\bibliographystyle{plain}
\bibliography{localBiblio1,transportation}

\end{document}